\documentclass[11pt,letterpaper]{article}
\usepackage{ijcnlp2017}
\usepackage{times}
\usepackage{latexsym}
\usepackage{caption}
\usepackage{hyperref}
\usepackage{url}
\usepackage{graphicx}
\usepackage{wrapfig}
\usepackage{amsmath}
\usepackage{amsfonts}
\usepackage{booktabs}
\usepackage{multirow}
\usepackage{multicol}
\usepackage{subfigure}
\usepackage{float}
\usepackage{nicefrac}       
\usepackage{microtype}      
\usepackage{siunitx}
\usepackage{caption}
\usepackage{subfigure}

\newcommand{\tabincell}[2]{\begin{tabular}{@{}#1@{}}#2\end{tabular}}

\makeatletter
\g@addto@macro\normalsize{%
  \abovedisplayskip 5pt plus 2pt minus 4pt%
  \belowdisplayskip \abovedisplayskip
  \abovedisplayshortskip 4pt plus3pt%
  \belowdisplayshortskip 4pt plus3pt minus3pt%
}
\makeatother

\ijcnlpfinalcopy

\title{DailyDialog: A Manually Labelled Multi-turn Dialogue Dataset}

\author{$^{\dagger}$Yanran Li\thanks{Authors contributed equally. Correspondence should be sent to Y. Li (csyli@comp.polyu.edu.hk).}, $^{\dagger,\mathsection}$Hui Su\footnotemark[1], $^{\ddagger}$Xiaoyu Shen, $^{\dagger}$Wenjie Li, $^{\dagger}$Ziqiang Cao, $^{\mathsection}$Shuzi Niu\\
       $^{\dagger}$Department of Computing, The Hong Kong Polytechnic University, Hong Kong\\
	  $^{\mathsection}$Institute of Software, Chinese Academy of Science, China\\
      $^{\ddagger}$Spoken Language Systems (LSV), Saarland University, Germany
       }

\date{}

\begin{document}

\maketitle

\begin{abstract}
We develop a high-quality multi-turn dialog dataset, \textbf{DailyDialog}, which is intriguing in several aspects. The language is human-written and less noisy. The dialogues in the dataset reflect our daily communication way and cover various topics about our daily life. We also manually label the developed dataset with communication intention and emotion information. Then, we evaluate existing approaches on DailyDialog dataset and hope it benefit the research field of dialog systems\footnote{The dataset is available on \url{http://yanran.li/dailydialog}}.
\end{abstract}

\section{Introduction}
Developing intelligent chatbots and dialog systems is of great significance to both commercial and academic camps. A good conversational agent enables enterprises to provide automatic customer services and thus reduce human labor costs. For academia, it is challenging yet appealing to build up such an intelligent chatbot which involves a series of high-level natural language processing techniques, such as understanding the underlying semantics of user input utterance, and generating coherent and meaningful responses. 

However, the training datasets for this research area are still deficient. Traditional dialogue systems are often trained with domain-specific spoken dialogue datasets~\cite{trains,dbox}, which are often small-scale and oriented to complete a specific task. More recent work feed their conversational models with open-domain datasets. Switchboard~\cite{switchboard} and OpenSubtitles~\cite{opensubtitles} datasets comprise approximately 150 turns in a ``conversation'' and thus are too disperse to capture the main topic. Twitter Dialog Corpus~\cite{twitter} and Chinese Weibo dataset~\cite{weibo} are comprised of posts and replies on social networks, which are noisy, informal and different from real conversations.
\begin{figure}
\fbox{\begin{minipage}{0.47\textwidth}
\textcolor{blue}{\textbf{A}}: I'm \textcolor{purple}{\underline{worried}} about something.\\
\textcolor{red}{\textbf{B}}: What's that?\\
\textcolor{blue}{\textbf{A}}: Well, I have to drive to school for a meeting this morning, and I'm going to end up getting stuck in rush-hour traffic.\\
\textcolor{red}{\textbf{B}}: That's \textcolor{purple}{annoying}, but nothing to worry about. \emph{Just breathe deeply when you feel yourself getting upset}.\\
\textcolor{blue}{\textbf{A}}: Ok, I'll try that.\\
\textcolor{red}{\textbf{B}}: Is there anything else \textcolor{purple}{\underline{bothering}} you?\\
\textcolor{blue}{\textbf{A}}: Just one more thing. A school called me this morning to see if I could teach a few classes this weekend and I don't know what to do.\\
\textcolor{red}{\textbf{B}}: Do you have any other plans this weekend?\\
\textcolor{blue}{\textbf{A}}: I'm supposed to work on a paper that'd due on Monday.\\
\textcolor{red}{\textbf{B}}: \emph{Try not to take on more than you can handle}.\\
\textcolor{blue}{\textbf{A}}: You're right. I probably should just work on my paper. \textcolor{purple}{\underline{Thanks}}!
\end{minipage}
}
\caption{An example in \textbf{DailyDialog} dataset. Some text is shortened for space. Best viewed in color.}
\label{fig:example}
\end{figure}

In this work, we develop a high-quality multi-turn dialogue dataset, which contains conversations about our daily life. We refer to it as \textbf{DailyDialog}. 
In our daily life, we communicate with others by two main reasons: \emph{exchanging information} and \emph{enhancing social bonding}. To exchange and share ideas, we often communicate with others following certain dialog flow. 
Typically, we do not rigidly answer others' questions and wait for the next question. Instead, humans often first respond to previous context and then propose their own questions and suggestions. In this way, people show their attention others' words and are willing to continue the conversation. Another reason why people communicate is to strengthen their social bonding with others. Therefore, daily conversations are rich in emotion. By expressing emotions, people show their mutual respect, empathy and understanding to each other, and thus improve the relationship between them.

We demonstrate the above two phenomena by an example conversation as in Figure~\ref{fig:example}. The \emph{words in Italic} are speaker B's own ideas that are new for the other speaker A. The \textcolor{purple}{\underline{underlined words in purple}} explicitly indicate the emotions. In the fourth speaker turn, speaker B first expresses his/her feeling on what he/she has heared from speaker A, which reveals his/her understanding. Then, speaker B suggests by saying \emph{Just breathe deeply when you feel yourself getting upset}. Following the direct response towards A, B's suggestion is original yet context-dependent. It shows that B builds up a connection link by responding to forgoing context and proposing new suggestions. 

We describe the dataset construction process and annotation criteria in Section~\ref{sec:dataset}, present and analyze the detailed characteristics in Section~\ref{sec:char}. We then evaluate existing mainstream approaches, including retrieval-based and generation-based approaches on the developed datasets in Section~\ref{sec:eval}. 

\section{Dataset Construction}
\label{sec:dataset}
\subsection{Basic Features and Statistics}
To construct a multi-turn dialog dataset, we crawl the raw data from various websites which serve for English learner to practice English dialog in daily life. That's why we refer it as \textbf{DailyDialog} dataset. The dialogues in the dataset preserve the following three appealing properties. 

First, the language in DailyDialog is human-written and thus is more formal than those datasets like Twitter Dialog Corpus~\cite{twitter} and Chinese Weibo dataset~\cite{weibo}. The latters are constructed by posts and replies on social networks, which are noisy, short and different from real conversations.

Second, the conversations in DailyDialog often focus on a certain topic and under a certain physical context. For example, a conversation happens in a shop is often between a customer looking for suitable goods and a salesman who is willing to help for purchasing. Another typical conversation happens between two students talking about their summer vacation trips. 

The third desirable feature is that the crawled dialogues usually end after reasonable speaker turns. This makes DailyDialog distinguished from existing dialog datasets such as Switchboard~\cite{switchboard} and OpenSubtitles~\cite{opensubtitles}, which often have 150+ and 1,000+ speaker turns in one ``conversation''. By examining some examples, we find that in such a conversation, people often talk about three or more topics (or scenes). Compared with them, our dataset has in average approximate 8 turns, which is more suitable to train compact conversational models.

After crawling, we de-duplicate the raw data, filter out those dialogues involving more than two parties (three or more speakers) and automatically correct the misspelling using autocorrect package\footnote{\url{https://github.com/phatpiglet/autocorrect/}}. Finally, the DailyDialog datasets contain 13,118 multi-turn dialogues. We also count the average speaker turns and tokens to give a brief view of the dataset. The resulting statistics are given in Table~\ref{basic}. From the statistics we can see, the speaker turns are roughly 8, and the average tokens per utterance is about 15. 

\begin{table}[h]
\centering
\begin{tabular}{rr}
\toprule
Total Dialogues & 13,118 \\
Average Speaker Turns Per Dialogue & 7.9 \\
Average Tokens Per Dialogue & 114.7 \\
Average Tokens Per Utterance & 14.6 \\
\bottomrule
\end{tabular}
\caption{Basic Statistics of DailyDialog.}
\label{basic}
\end{table}

\begin{figure*}[htbp]
\centering 
\subfigure[Emotion distributions in DailyDialog.]{ 
\begin{minipage}{5cm}
\centering 
\includegraphics[height=1.2in]{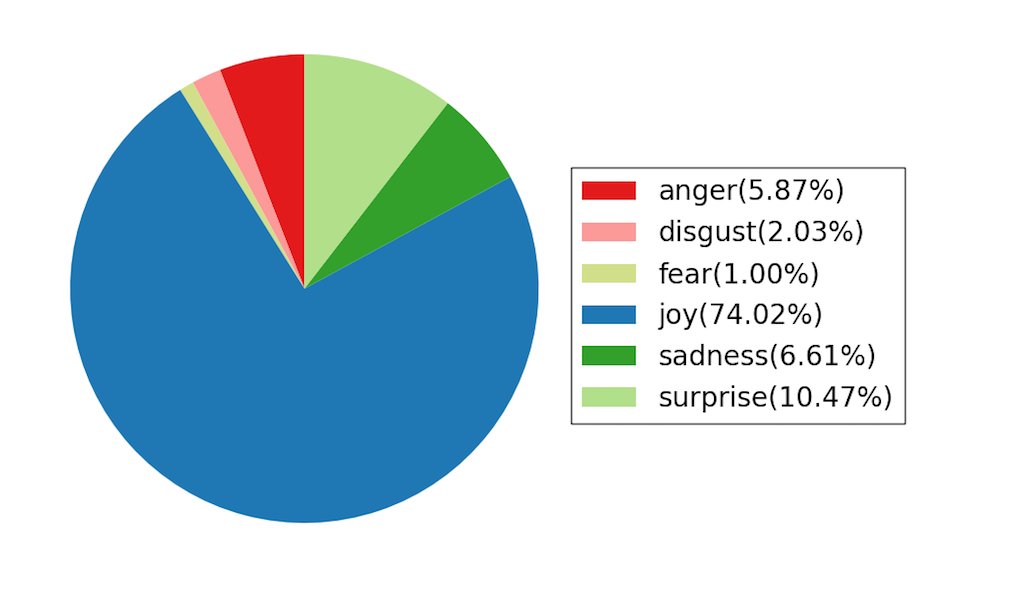}
\label{fig:a}
\end{minipage}
}
\subfigure[Topic distributions in DailyDialog.]{ 
\begin{minipage}{5cm}
\centering 
\includegraphics[height=1.2in]{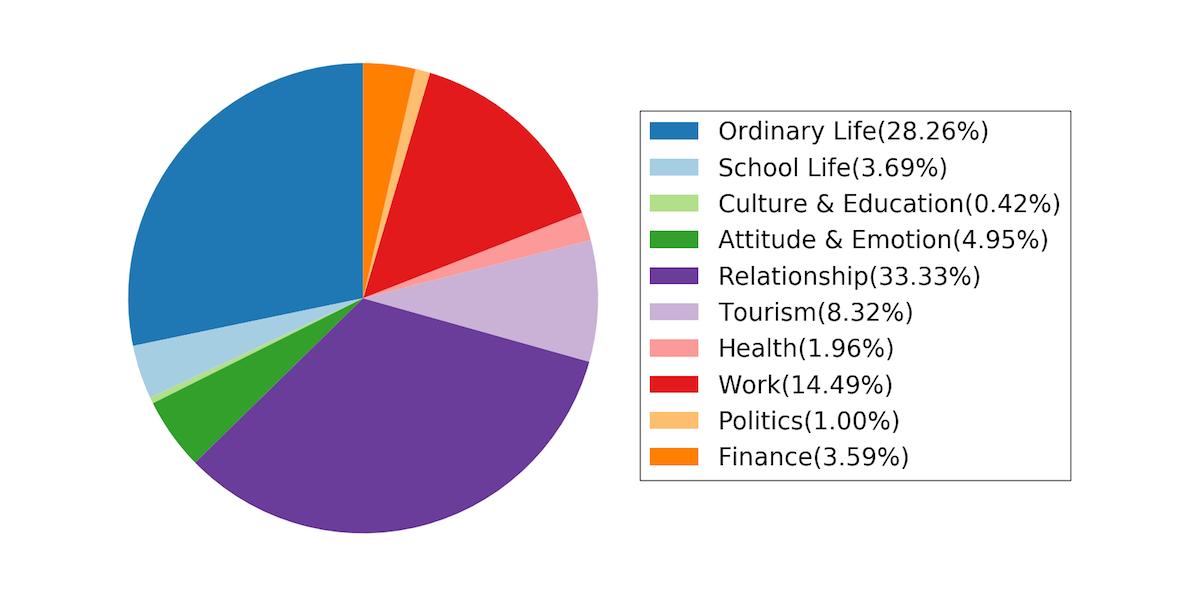} 
\label{fig:b}
\end{minipage}
}
\subfigure[Interactions of dialog acts in each utterance pairs.]{ 
\begin{minipage}{5cm}
\centering 
\includegraphics[height=1.35in,width=1.7in]{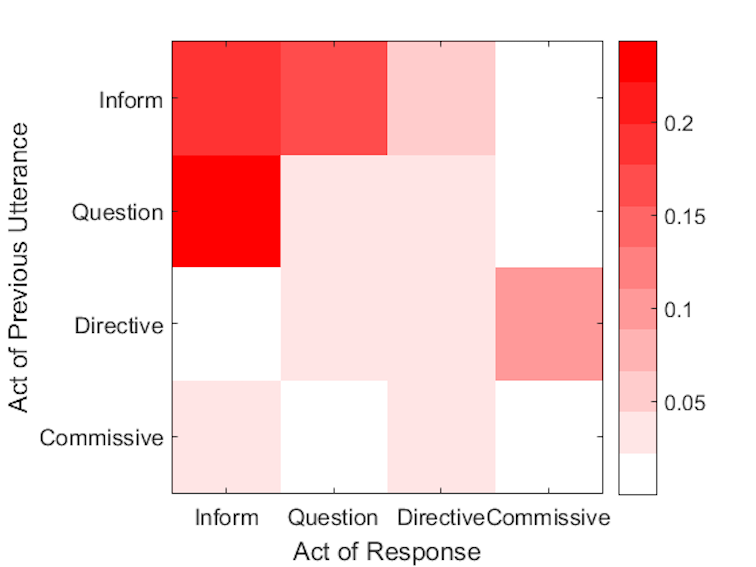} 
\label{fig:c}
\end{minipage}
}
\caption{Statistics in DailyDialog.} 
\label{fig:2} 
\end{figure*}

\subsection{Annotation Criteria and Procedure}
\label{sec:label}
Because the dialogues in DailyDialog datasets are written to reflect our daily conversations, they mainly conform certain communication ways. As stated before, the purpose of the dialogues are \emph{exchanging information} and \emph{enhancing social bonding}. To allow further research on our daily communication behaviors, we manually label the DailyDialog dataset to reflect the two purposes. 

The communication purpose of exchanging information is related to the communication intentions. This factor has been extensively explored under the name of dialog act and speech act. In general, dialog acts represent the communication functions when people saying something. To label the dialog acts in DailyDialog, we follow the criteria in~\citet{Amanova2016CreatingAD} because it is adaptive to mainstream annotation criteria ISO 24617-2~\cite{criteria} and consistent with existing annotated dataset such as Trains~\cite{trains} and DBox~\cite{dbox}. Following~\citet{Amanova2016CreatingAD}, we label each utterance as one of four dialog act classes: \{Inform, Questions, Directives, Commissive\}. The \textbf{Inform} class contains all statements and questions by which the speaker is providing information. The \textbf{Questions} class is labeled when the speaker wants to know something and seeks for some information. The \textbf{Directives} class contains dialog acts like request, instruct, suggest and accept/reject offer. The \textbf{Commissive} class is about accept/reject request or suggestion and offer. The former two classes are information transfer acts, while the latter two are action discussion acts. Detailed explanations can be found in~\citet{Amanova2016CreatingAD}. Thereafter, in the DailyDialog dataset, we have four intention classes. 

The second communication purpose, enhancing social bonding, is highly correlated with human emotion. Following~\cite{weibo}, we adopt the ``BigSix Theory''~\cite{bigsix} to label each utterance in DailyDialog. \citet{bigsix} thinks that there are six primary and universal emotions in human beings: \{Anger, Disgust, Fear, Happiness, Sadness, Surprise\}. Besides the main six categories of emotions, we find it necessary to add additional category to represent other emotions. Hence, we have seven emotion categories in DailyDialog. 

To guarantee the annotation quality, we recruit three experts who have good knowledge in dialog and communication theory. After teaching them the criteria, we sample 100 dialogues for them to annotate and reduce the discrepancy by discussion among them. Then, they independently annotate the whole dataset and achieve the inter annotator agreement of 78.9\%. When the disagreement happens, we follow the majority rule or let them re-annotate to find a ``common'' annotation. The detailed statistics of the final annotation information are given in the following section.

\section{Characteristics}
\label{sec:char}
In this section, we delve deeply into DailyDialog datasets, and show our datasets are beneficial in several aspects:
\begin{itemize}
\item \emph{Daily Topics}: It covers ten categories ranging from ordinary life to financial topics, which is different from domain-specific datasets.
\item \emph{Bi-turn Dialog Flow}: It conforms basic dialog act flows, such as Questions-Inform and Directives-Commissives bi-turn flows, making it different from question answering (QA) datasets and post-reply datasets.
\item \emph{Certain Communication Pattern}: It follows unique multi-turn dialog flow patterns reflecting human communication style, which are rarely seen in task-oriented datasets.
\item \emph{Rich Emotion}: It contains rich emotions and is labeled manually to keep high-quality, which is distinguished from most existing dialog datasets.
\end{itemize}

\subsection{Daily Topics}
The dialogues in the developed dataset happens in our everyday life, and that's why we name it \textbf{DailyDialog}. They cover a wide range of daily scenarios: chit-chats about holidays and tourisms, service-dialog in shops and restaurants, and so on. After looking into its topics, we cluster them into ten categories. The statistics for each category is summarized in Figure~\ref{fig:b}.

The largest three categories are: Relationship (33.33\%), Ordinary Life (28.26\%) and Work (14.49\%). This is also consistent with our real experience that we often invite people for social activities (Relationship), talk about what happened recently (Ordinary Life) and what happened at work (Work). 

\subsection{Bi-turn Dialog Flow}
Because the dialogues are assumed to happen in daily life, they follow natural dialog flow. It makes DailyDialog dataset quite different from existing QA datasets such as SubTle dataset~\cite{moviedialog}，which are improperly used for training dialog systems. DailyDialog dataset also distinguishes from those post-reply datasets such as Reddit comment~\cite{reddit}, Sina Weibo~\cite{Shang2015NeuralRM} and Twitter~\cite{twitter} datasets. The latter datasets comprise post-reply pairs on social networks where people interact with others more freely (often more than two speakers) and results in ambiguous dialog flows. 

Instead, the dialog act flows in Dailydialog are more consistent with our daily communication. For example, we usually do not leave alone others' question and just tersely change the topic. Instead, we will answer others' questions politely. By the definitions we introduce in Section~\ref{sec:label}, this reflects a \emph{Questions-Inform} bi-turn dialog flow. This is a frequent circle phenomena because it represents a information transfer between the two speakers in the dialog. Another example is that when someone proposes a idea, such as going out for dinner, the other speaker in the dialog usually responds to this proposal. This reflects a \emph{Directives-Commissives} dialog flow and captures the speakers' suggestions and commitments to conduct certain acts. By labeling each utterances in dialogues, Dailydialog datasets contain more than ten thousands examples of approximately 8-turn dialog act flows. We hope this is beneficial for the research in dialog management. The distributions of these four dialog acts are given in Table~\ref{intention}. We also demonstrate the interactions between each four dialog acts in Figure~\ref{fig:c}.

\begin{table}[h]
\centering
\begin{tabular}{cccc}
\toprule
Inform & Questions & Directives & Commissive \\
\midrule
46,532&29,428&17,295&9,724 \\
45.2\%&28.6\%& 16.8\%&9.4\% \\
\bottomrule
\end{tabular}
\caption{Intention Statistics in DailyDialog.}
\label{intention}
\end{table}

\subsection{Certain Communication Pattern}
Besides the basic \emph{Questions-Inform} and \emph{Directives-Commissives} bi-turn dialog flows, we also find two unique multi-turn flow patterns in DailyDialog dataset. 

\noindent{\textbf{Pattern 1:}} In human-to-human communication, people are inclined to both answer the questions and then initiate a new question to let the dialog last. In other words, a speaker can change from information-provider to information-seeker in a single speaker turn. We find 2,398 (18.3\%) dialogues in DailyDialog exhibits this patterns, which is quite frequent.

\noindent{\textbf{Pattern 2:}} When someone is proposing an activity or offering a suggestion, the other speaker usually comes up with another idea. This is sensible because the two speakers often have different views about a topic and by exchanging different proposals, they persuade and influence the other. This results in a \emph{Directives-Directives-Commissives}-like pattern in dialog flows, which happens totally 1,203 times (9.2\%) in our dataset. 

The two patterns shed light on our daily communications style, which are merely found in single-turn datasets or task-oriented datasets like Ubuntu~\cite{ubuntu} and restaurant reservation datasets~\cite{babi}.

\subsection{Rich Emotion}
As discussed before, the other main purpose of our daily communication is \emph{enhancing social bonding}. Hence, people tend to express their emotions during communication. When hearing from others' miseries, we often say ``I'm sorry to hear that'' or ``What a poor guy''. And when we appease others, the listener often feels better. Such emotional words are rich in DailyDialog dataset. Because automatic emotion classification is difficult~\cite{ECM}, we manually label the emotion for each utterance to make them as accurate as possible. This distinguishes DailyDialog datasets from most existing dialog datasets. Similarly, we summarize the basic statistics on labelled emotion in Table~\ref{emotion}.\footnote{The imbalanced emotion categories suggest that it might be improper to label the emotion following ``BigSix'' Theory~\cite{bigsix}. However, we keep it in this work to follow previous work~\cite{weibo}. To propose a novel emotion theory is beyond this work.} 

\begin{table}[h]
\centering
\begin{tabular}{rccc}
\toprule
& Count & of EU & of Total \\
\midrule
Anger&1022  &5.87  & 0.99 \\
Disgust& 353 & 2.03&   0.34\\
Fear & 74 & 1.00 &  0.17\\
Happiness& 12885&  74.02   &  12.51\\
Sadness& 1150 & 6.61  & 1.12\\
Surpise& 1823& 10.47& 1.77\\
Other& 85572&-& 83.10\\
\bottomrule
\end{tabular}
\caption{Emotion Statistics in DailyDialog. EU denotes for utterances that contain the main six categories of emotion, while Total denotes for all utterances in the dataset. Numbers are multiplied by 100\%.}
\label{emotion}
\end{table}

Additionally, we observe in our daily life, a healthy and pleasant conversation often ends with positive emotions. Therefore we examine our DailyDialog dataset by how many conversations are ending or positive emotions (i.e., happy), and find 3,675 (28.0\%) ``happy'' dialogues. We also count how many conversations have changed to positive emotions even though they begin with negative emotions (e.g., sad, disgust, anger) and find 113 (0.8\%) such examples. We hope our dataset facilitates future research on developing conversational agents able to regulate the conversation towards a happy ending. 


\section{Evaluating Existing Approaches}
\label{sec:eval}
In this section, we evaluate existing mainstream approaches on the proposed DailyDialog. We mainly compare five categories of approaches: (1) Embedding-based Similarity for Response Retrieval~\cite{luocombination}; (2) Feature-based Similarity for Response Retrieval~\cite{jafarpour2010filter}; (3) Feature-based Similarity for Response Retrieval and Reranking~\cite{luocombination,otsuka2017utterance}; (4) Neural network-based for Response Generation~\cite{Shang2015NeuralRM,hred}; (5) Neural network-based for Response Generation with Labeling Information~\cite{ECM}. All the evaluated approaches are implemented by TensorFlow~\cite{tensorflow}.

\begin{table*}[h]
\centering
\begin{tabular}{rccccccc}
\toprule
& Epoch & Test Loss & PPL & BLEU-1 & BLEU-2 & BLEU-3 & BLEU-4 \\
\midrule
Seq2Seq & 30 &  4.024 & 55.94 & 0.352 & 0.146 & 0.017 &0.006 \\
Attn-Seq2Seq & 60 & 4.036 & 56.59 &0.335& 0.134& 0.013& 0.006 \\
HRED & 44 & 4.082 & 59.24 &0.396& 0.174& 0.019& 0.009\\
\hline
L+Seq2Seq & 21 & 3.911 & 49.96&0.379&0.156&0.018&0.006\\
L+Attn-Seq2Seq & 37 & 3.913 & 50.03&0.464& 0.220& 0.016& 0.009\\
L+HRED & 27 & 3.990 & 54.05&0.431& 0.193& 0.016& 0.009\\
\hline
Pre+Seq2Seq & 18 & 3.556 & 35.01&0.312&0.120&0.0136&0.005\\
Pre+Attn-Seq2Seq & 15 & 3.567&35.42&0.354&0.136&0.013&0.004\\
Pre+HRED & 10 &3.628&37.65&0.153& 0.026&0.001&0.000\\
\bottomrule
\end{tabular}
\caption{Experiments Results of generation-based approaches.}
\label{generation_ppl}
\end{table*}

\subsection{Experimental Setup}
\label{sec:setup}
We randomly separate the DailyDialog datasets into training/validation/test sets with 11,118/1,000/1,000 conversations. We tune the parameters on validation set and report the performance on test sets. In all experiments, the vocabulary size is set as 25,000 and all the OOV words are mapped to a special token UNK. We set word embeddings to size of 300 and initialize them with Word2Vec embeddings trained on the Google News Corpus\footnote{\url{ttps://code.google.com/archive/p/word2vec/}}. The encoder and decoder RNN in the following experiments are 1-layer GRU with 512 hidden neurons~\cite{gru}. All the trained model parameters are then used as an initialization point. We set the batch size as 128 and fix the learning rate as 0.0002. Models are trained to minimize the cross entropy using Adam optimizer~\cite{kingma2014adam}.

\subsection{Retrieval-based Approaches}
\subsubsection{Compared Approaches}
First, we choose three categories of four retrieval-based approaches, i.e., (1) Embedding-based Similarity~\cite{luocombination}; (2) Feature-based Similarity~\cite{jafarpour2010filter,docchat}; (3)(4) Feature-based Similarity with Intention and Emotion Reranking~\cite{luocombination,otsuka2017utterance}. We aim to see whether classical embeddings-based, feature-based and reranking-enhanced approaches are effective on DailyDialog.

\noindent{\bf{Embedding-based}} The embedding-based approach is using basic neural networks as described in Section~\ref{sec:setup} and denoted as \{Embedding\} below. We measure the distance between embeddings as the average of cosine similarity, Jaccard distance and Euclidean distance. At test time, candidates whose context embedding is closer to the test context embedding are ranked higher. Similar approaches have been adopted extensively on response retrieval task, such as~\citet{luocombination}.

\noindent{\bf{Feature-based}} We then evaluate the performance of feature-based retrieval approach. We adopt several linguistic features: TF-IDF and three fuzzy string matching features, i.e., QRatio, WRatio, and Partial ratio. We first use TF-IDF to select 1,000 candidates and rank them with the fuzzy features. These fuzzy features is implemented with fuzzywuzzy package\footnote{\url{https://github.com/seatgeek/fuzzywuzzy}}. We denote this feature engineering approach as \{Feature\}. Similar approaches have been demonstrated effectively on response retrieval task and duplicate question detection task\footnote{\url{https://github.com/abhishekkrthakur/is_that_a_duplicate_quora_question}}, such as~\citet{docchat,luocombination}.

\begin{table}[h]
\centering
\begin{tabular}{rcccc}
\toprule
& BLEU-2 & BLEU-3 & BLEU-4 \\
\midrule
Embedding & 0.207&0.162& 0.150 \\\cline{2-4}
\bf{Feature}& \bf{0.258}&\bf{0.204}& \bf{0.194} \\
+ I-Rerank & 0.204&0.189&0.181 \\
+ I-E-Rerank&0.190&0.174& 0.164 \\
\bottomrule
\end{tabular}
\caption{BLEU scores of retrieval-based approaches.}
\label{retrieval_bleu}
\end{table}

\noindent{\bf{Reranking By Intention}} We also examine reranking-enhanced retrieval approaches, which encourages the retrieved response to follow a certain rules. \citet{luocombination} provides a simplest way to realize it. Because intention has shown as a beneficial factor in response selection~\cite{otsuka2017utterance}, we first examine reranking-enhanced retrieval approach based on the intention (dialog act) label in DailyDialog dataset. We compare the intention history of the test example with that of the candidate example, and use the compared similarity as reranking feature. For example, if the test intention history is \{2,1,3\}, then the candidate response whose intention history is also \{2,1,3\} will be reranked higher. Based on the feature-based retrieval approach, we denote the reranking by intention as \{+I-Rerank\}.

\noindent{\bf{Reranking By Intention \& Emotion}} The last retrieved-based approach we evaluate is similar with \{+I-Rerank\}, with the only difference that the candidate responses are reranked by both intention and emotion labels. We denote it as \{+I-E-Rerank\}.

Because the groundtruth responses in the test set are not seen in the training set, we can not evaluate the performance using ranking-like metrics such as Recall-k. We instead report the BLEU scores achieved by retrieval-based approaches in Table~\ref{retrieval_bleu}. 

We also evaluate them by calculating the ``Equivalence'' percentage between the labels (i.e., intention, emotion) of the retrieved responses and those of the groundtruth responses. The results are reported in Table~\ref{reranking_acc}. Though subtle improvements can be seen when using labels, we speculate it as not a very strong evaluation metric. It is unsafe to conclude that the higher the ``equivalence'' percentage is, the better (more coherent, more suitable) the retrieved response will be. 

\begin{table}[h]
\centering
\begin{tabular}{rcccc}
\toprule
& Feature & +I-Rerank & +I-E-Rerank \\
\midrule
Intention  &46.3 & \bf{47.3}& 46.7 \\
Emotion & 73.7 & 72.3 & \bf{74.3}\\
\bottomrule
\end{tabular}
\caption{``Equivalence'' percentage (\%) of retrieval-based approaches.}
\label{reranking_acc}
\end{table}

\subsubsection{Intention And Emotion Matters}
In dialog response generation, word-level overlap metrics such as BLEU are inadequate~\cite{hownot}. To provide insights on whether intention and emotion are beneficial, and how they works, we conduct several case studies in Table~\ref{case:retrieval}.

\begin{table*}[h]
  \centering
  \renewcommand{\arraystretch}{0.9}
  \resizebox{\textwidth}{!}{
  \begin{tabular}{p{0.4\textwidth}p{0.47\textwidth}}
    \toprule
     Test Context  &  Retrieved Response \\
    \midrule
    \tabincell{l}{U1: \emph{Can you direct me to Holiday inn ?} (3) \\ U2: \emph{Cross the street... You can't miss it.} (3)\\  GA: \textbf{Thanks.}}  & \tabincell{l}{F: \emph{Well, we've got some great mangoes on sale.}\\ +I: \emph{About how long will it take me to get there?}\\ +I-E: \emph{About how long will it take me to get there?}} \\\midrule
    \tabincell{l}{U1: \emph{No way... You can't keep it here.} (1) \\ U2: \emph{Please...it's so cute and tame.} (0)\\ U3: \emph{All right. But you have to...} (0) \\ GA: \textbf{I will. Thank you, Mummy.}} & \tabincell{l}{F: \emph{Is there somewhere you wanted to go eat at?}\\ +I: \emph{Sprite with ice, please.}\\ +I-E: \emph{Now we get along very well. It makes me feel...}} \\
    \bottomrule
  \end{tabular}
  }
  \caption{Case Study of Reranking-enhanced Retrieve Approaches. Context words are shortened for space.}
  \label{case:retrieval}
\end{table*}

\begin{table*}[h]
  \centering
  \renewcommand{\arraystretch}{0.9}
  \resizebox{\textwidth}{!}{
  \begin{tabular}{p{0.4\textwidth}p{0.47\textwidth}}
    \toprule
     Test Context  &  Generated Response \\
    \midrule
    \tabincell{l}{U1: \emph{I have to check out today.} \\ ~~~~~~~\emph{I'd like my bill ready by 10 in morning.} \\ U2: \emph{You can be sure of that, sir .}\\  GA: \textbf{Thank you.}}  & \tabincell{l}{Attn: \emph{all right, sir.}\\ Pre+Attn: \emph{how long will it take to get there?}\\ HRED: \emph{here you are.} \\ Pre+HRED: \emph{how long will it take to get there?}} \\
    \bottomrule
  \end{tabular}
  }
  \caption{Case Study of Generation-based Approaches.}
  \label{case:generation}
\end{table*}

In the first block, we give a example of how intention helps to find more proper response. The intentions in the test context (U1 \& U2) are \{3, 3\}, meaning \{Directives, Directives\}. The gold answer (GA) in the test set is ``Thanks.'' Although both three retrieved responses are not exactly same with GA, the approaches that reranking by intention (+I) and reranking by intention and emotion (+I-E) find more suitable response than the feature-based approach without reranking (F). It is because, the context corresponding to the retrieved response ``About how long will it take me to get there?'' is ``Excuse me, but can you tell me the way... Just go straight... You can ’ t miss it'', whose dialog act flow \{3, 3\} is consistent with the context test. On the contrary, the response found by feature-based approach has the context ``Can you direct me to some fresh produce that's on sale?'', which should be attributed to the poor result.

Similar cases are given in the second block where emotion history information benefits. The emotions in the test context (U1, U2 \& U3) are \{1, 0, 0\}, meaning \{Anger, Others, Others\}. The most proper retrieved responses are from the reranking approach by intention and emotion (+I-E) that finds ``Now we get along very well. It makes me feel that I'm someone special. It makes me feel that I'm someone special.'' The context history for this response is ``oh, really? so you just took home a stray cat? // Yes. It was starving and looking for something to eat when I saw it. // Poor cat.'' whose emotion history is \{6, 0, 0\}.

\subsection{Generation-based Approaches}
\subsubsection{Compared Approaches}
\noindent{\bf{Seq2Seq}} The simplest generation-based approach we adopt is a vanilla Seq2Seq with GRU as basic cell, as described in Section~\ref{sec:setup}. Such approach is widely selected as baseline models in dialog generation~\citet{Shang2015NeuralRM,ubuntu,reddit}.

\noindent{\bf{Attention-based Seq2Seq}} We then evaluate the Seq2Seq approach with attention mechanism~\cite{attention} which has shown its effectiveness on various NLP tasks including dialog response~\cite{teaching,Luong2015EffectiveAT,Mei2017CoherentDW}. We denote this approach as \{Attn-Seq2Seq\}.

\noindent{\bf{HRED}} The third generation-based approach we evaluate is hierarchical encoder-decoder (HRED)~\cite{hred}. Due to its context-aware modeling ability, HRED has shown better performances in previous work~\cite{hred}.

\noindent{\bf{Intention and Emotion-enhanced}} To utilize the intention and emotion labels, we follow~\citet{ECM} to incorporate the label information during decoding. The intention and emotion labels are characterized as one-hot vectors. We denote the label-enhanced approaches as \{L+\} and the performances are given in the second box in Table~\ref{generation_ppl}.

\noindent{\bf{Pretrained}} We also examine whether pre-training with other dataset will boost the performance of the first three generation-based approaches. Following~\citet{Li2016DeepRL,li2017adversarial}, we use the OpenSubtitle dataset~\cite{opensubtitles}\footnote{\url{https://github.com/jiweil/Neural-Dialogue-Generation}}. Because it has no clear and concise segmentation for each conversation, we treat each of three consecutive utterances as context, and the foregoing one as response. Finally, 3,000,000 three-turn dialogs are randomly sampled and used to pre-train the compared models for 12 epochs. We denote the approaches using pre-training as \{Pre+\}.

According to BLEU scores from Table~\ref{generation_ppl} (last four columns), we can see that in general attention-based approaches are better than vanilla Seq2Seq model. Among the three compared approaches, HREDs achieve highest BLEU scores because they take history information into consideration. Furthermore, label information is effective even though we utilize them in the simplest way. These findings are consistent with previous work~\cite{hred,vhred}. 

On the other hand, the first three columns in Table~\ref{generation_ppl} show that models pretrained by OpenSubtitle converge faster, achieving lower Perplexity (PPL) but poorer BLEU scores. We conjecture it as a result of domain difference. OpenSubtitle dataset is constructed by movie lines, whereas our datasets are daily dialogues. Moreover, OpenSubtitles has approximately 1000+ speaker turns in one “conversation”, while our dataset has in average 8 turns. To pretrain a model by corpus from different domain will harm its performance on the target domain. Hence, it is less optimal to simply pretrain models with large-scale datasets such as OpenSubtitle, which is domain different from the evaluation datasets. We further examine this issue by comparing the generated answers by models trained solely on DailyDialog with and without pre-training. 

\subsubsection{Case Study}
We give a case study in Table~\ref{case:generation}. It can be seen the two pre-trained models (the second and the fourth row) generate responses that are irrelevant with the context. In contrast, the corresponding model without pre-training produce more reasonable responses.

\section{Related Work}
\subsection{Domain-Specific Datasets}
The research on chatbots and dialog systems is still new and developing. Literature on traditional dialog system primarily relies on template-based and retrieval-based approaches and applies to specific-domain of data. 

Popular datasets for this research area include TRAINS~\cite{trains}, DBOX~\cite{dbox}, bAbI synthetic dialog~\cite{babi} and Movie Dialog datasets~\cite{moviedialog}. These datasets feature different types of dialogues happening in different physical contexts. For example, the TRAINS corpus contains problem-solving dialogues and the dialog systems trained with TRAINS are performing as task-oriented assistants. The tasks are often about the shipping of railroad goods and thereafter it is called TRAINS. The bAbI~\cite{babi} and Movie Dialog dataset~\cite{moviedialog} contain dialogues about movies and the tasks in these datasets are movie question answering, movie recommendation and so on. Another popular dataset is Ubuntu dataset~\cite{ubuntu} which extracts the user posts and replies in Ubuntu forums and the task is to answer users' computer-related questions.

\subsection{Open-Domain Datasets}
More recent work concentrates on generation-based approaches, which are mainly based on the sequence-to-sequence encoder-decoder architecture~\cite{hred,vhred}. These generation-based approaches are often trained with large-scale open-domain datasets. 

In~\citet{Shang2015NeuralRM}, the authors propose a neural responding machine (NRM) and examine their approach on Sina Weibo dataset~\cite{weibo}. The Sina Weibo dataset is constructed by crawling users' posts and replies on a Chinese social network. Similar dataset is constructed by~\citet{twitter} who provides a Twitter dataset. Besides social network,~\citet{reddit} constructs a dialog training dataset with Reddit Forum posts. Existing work based on neural networks has examined their approaches on these datasets~\cite{ECM,weibo,hred,vhred}. Although these datasets are large-scale, the dialogues in them are often noisy and short. Even worse, the artificially constructed post-reply pairs are different from our real conversations. 

To train neural network based conversational models, researchers often pre-train their models by using movie subtitles which are large-scale and conversation-like. The most widely adopted dataset is OpenSubtitle~\cite{opensubtitles} which is used in ~\citet{Li2016DeepRL,li2017adversarial}. Other similar datasets are SubTle dataset~\cite{babi} which is then used to build up MovieQA sub-dataset and MovieTriples~\cite{movietriples}.

\section{Conclusions and Future Work}

In this work, we develop the dataset DailyDialog which is high-quality, multi-turn and manually labeled. We show the proposed dataset is appealing in four main aspects. The dialogues in the dataset cover totally ten topics and conform common dialog flows such as Questions-Inform and Directives-Commissives bi-turn flows. In addition, DailyDialog contains unique multi-turn dialog flow patterns, which reflect our realistic communication ways. And it is rich in emotion. The evaluation results in Section~\ref{sec:eval} are initial but indicative. 

In the future we plan to design advanced mechanisms to explore the unique multi-turn dialog flows described in Section~\ref{sec:char}. It is also promising to utilize the topic information in our dataset by domain adaptation and transfer learning. Our dataset is available on {\url{http://yanran.li/dailydialog}}, and we hope it is beneficial for future research in this field. 

\section*{Acknowledgements}
We appreciate for the valuable suggestions and comments from the anonymous reviewers. The work described in this paper was supported by Research Grants Council of Hong Kong (PolyU 152094/14E, 152036/17E), National Natural Science Foundation of China (61672445, 61272291 and 61602451) and The Hong Kong Polytechnic University (GYBP6, 4-BCB5, B-Q46C).

\bibliography{ijcnlp2017}
\bibliographystyle{ijcnlp2017}

\end{document}